\pdfoutput=1

\documentclass[11pt]{article}

\usepackage[final]{acl}

\usepackage{times}
\usepackage{latexsym}

\usepackage[T1]{fontenc}

\usepackage[utf8]{inputenc}

\usepackage{microtype}

\usepackage{inconsolata}

\usepackage{graphicx}

\usepackage{hyperref}
\usepackage{url}
\usepackage{xcolor,colortbl}
\usepackage{tcolorbox}
\usepackage{booktabs}
\usepackage{graphicx}
\usepackage{mathtools}
\usepackage{lineno}
\usepackage{multirow}
\usepackage{amsmath}
\usepackage{amsthm}
\usepackage{amssymb}
\definecolor{darkblue}{rgb}{0, 0, 0.5}
\hypersetup{colorlinks=true, citecolor=darkblue, linkcolor=darkblue, urlcolor=darkblue}

\DeclarePairedDelimiter{\norm}{\lVert}{\rVert}

\newcommand{\Cov}{{\rm Cov}}

\newcommand{\fs}{{\mathbf s}}
\newcommand{\fx}{{\mathbf x}}
\newcommand{\fmu}{\boldsymbol{\mu}}
\newcommand{\fW}{{\mathbf W}}
\newcommand{\fI}{{\mathbf I}}
\newcommand{\bR}{{\mathbb R}}
\newcommand{\cX}{{\mathcal X}}
\newcommand{\cM}{{\mathcal{B}}^{+}}
\newcommand{\Var}{\mathrm{Var}}
\newcommand{\cS}{{\mathcal S}}

\newcommand{\cB}{{\mathcal B}}
\newcommand{\E}{\mathrm{E}}
\newcommand{\vect}{\mathrm{vec}}
\newcommand{\tr}{\mathrm{tr}}
\newcommand{\dd}{\mathrm{d}}

\newtheorem*{definition}{Definition}
\newtheorem{proposition}{Proposition}
\title{BBScoreV2: Learning Time-Evolution and Latent Alignment from Stochastic Representation }

\author{%
  Tianhao Zhang\textsuperscript{1}\thanks{ Equal contribution.}, 
  Zhecheng Sheng\textsuperscript{1}\footnotemark[1], 
  Zhexiao Lin\textsuperscript{2}\footnotemark[1], 
  Chen Jiang\textsuperscript{1}\footnotemark[1], 
  Dongyeop Kang\textsuperscript{1} \\
  \textsuperscript{1}University of Minnesota, Twin Cities \\
  \textsuperscript{2}University of California, Berkeley \\
  \texttt{\{zhan7594, sheng136, jian0649, dongyeop\}@umn.edu}, 
  \texttt{zhexiaolin@berkeley.edu}
}

\begin{document}

\maketitle

\begin{abstract}
Autoregressive generative models play a key role in various language tasks, especially for modeling and evaluating long text sequences. While recent methods leverage stochastic representations to better capture sequence dynamics, encoding both temporal and structural dependencies and utilizing such information for evaluation remains challenging. In this work, we observe that fitting transformer-based model embeddings into a stochastic process yields ordered latent representations from originally unordered model outputs. Building on this insight and prior work, we theoretically introduce a novel likelihood-based evaluation metric BBScoreV2. Empirically, we demonstrate that the stochastic latent space induces a "clustered-to-temporal ordered" mapping of language model representations in high-dimensional space, offering both intuitive and quantitative support for the effectiveness of BBScoreV2. Furthermore, this structure aligns with intrinsic properties of natural language and enhances performance on tasks such as temporal consistency evaluation (e.g., Shuffle tasks) and AI-generated content detection.
\end{abstract}

\section{Introduction}
Generative models are rapidly gaining traction in NLP \citep{zou2023survey, yang2023diffusion, yi2024diffusion}, particularly for the complex task of modeling and generating long text sequences—a challenge central to downstream applications such as text generation and machine translation. Recently, stochastic representations of latent spaces have emerged as a promising approach, showing considerable success in areas including time-series analysis \citep{Liu:2021}, dynamical flow modeling \citep{Albergo:2023, Albergo:2023BNF}, and video generation \citep{Zhang:2023}. In the context of text generation, \citet{Wang:2022} introduced a method that models long sequences as stochastic dynamical flows, yielding strong results in producing coherent long texts. However, accurately learning the time-dependent probability density functions inherent in text data remains an open problem. Furthermore, effectively leveraging the information encoded in stochastic representations continues to be a significant challenge that has not yet been fully addressed.

\paragraph{Brownian bridge (BB) process helps to learn time-evolution in the stochastic representation} While the temporal evolution captured in articles offers insights into linguistic properties like coherence and theme \citep{Sheng:2024}, effectively encoding this temporal information into latent representations remains difficult. Drawing inspiration from the Time-control model \citep{Wang:2022} and Stochastic Interpolation \citep{Albergo:2023BNF,Albergo:2023}, we propose using the "bridge process" from stochastic process theory \citep{Oksendal:2003} to \textbf{encode and evaluate} sentence-level temporal information within latent representations. Furthermore, by leveraging the raw embeddings from frozen language models, we can also incorporate sentence-level structural information. In this work, we utilize the BB, the simplest bridge process characterized by fixed start and end points \citep{Oksendal:2003} and widely applied across various domains. We believe that more complex bridge processes, such as the Schrödinger bridge \citep{Albergo:2023BNF,Albergo:2023}, could offer richer encoding capabilities, representing a promising avenue for future research.

\begin{figure*}[ht!]
\centering
\includegraphics[width=0.9\linewidth]{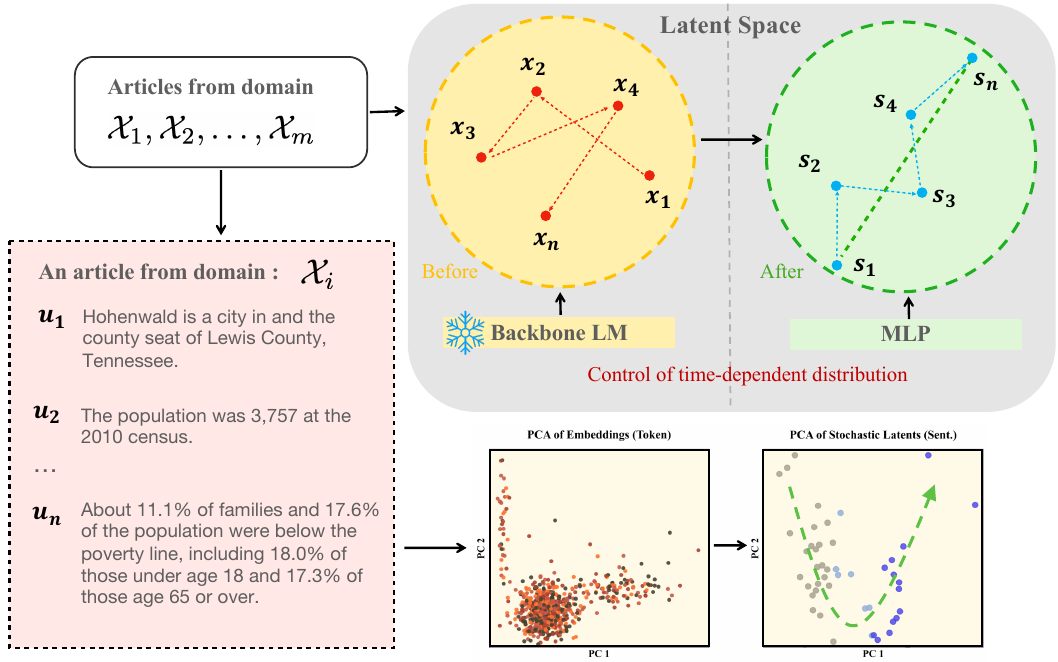}
\caption{Schematic diagram of the Stochastic Representation in the latent space. An article from domain $\cX_i$,  segmented into sentences $(u_1, u_2, \cdots u_n)$, is processed by the encoder which consists of a pre-trained language model (LM) and a multi-layer perceptron (MLP). The encoder maps each sentence into latent space and after optimizing for the stochastic objective, the latent trajectory becomes time dependent.} 
\label{fig:method_distribution_learning}
\end{figure*}

To evaluate such encoded time-evolution information, we introduce BBScoreV2, a novel likelihood-based evaluation metric for long-text assessment. BBScoreV2 evaluates the time evolution within a stochastic representation by considering both its temporal and structural dependencies, as detailed in Section \ref{sec:sr}. This metric is particularly useful for article coherence evaluation, exemplified by the shuffle task, which disrupts the natural temporal order. Existing methods for this task \citep{lai-tetreault-2018-discourse, Jeon:2022} often depend heavily on the training domain, are limited by their training paradigms, can only assess pairwise data, and are restricted to articles of the same length. In contrast, BBScoreV2 assesses general temporal order, offering greater flexibility while maintaining comparable performance. To demonstrate this, we generalize the standard Shuffle test \citep{Barzilay:2008, joty-etal-2018-coherence, Moon:2019} into a more robust Mixed Shuffle test. This new test compares shuffled and unshuffled versions both within and between different articles, allowing for evaluation of the metric's robustness independent of individual article characteristics like length. Furthermore, BBScoreV2 proves valuable in downstream applications such as Human-AI discrimination and exhibits strong performance in out-of-domain (O.O.D.) scenarios, likely due to its ability to capture the general structural and temporal information in human writing and preserve it in the stochastic representation.

The main contributions of our work can be summarized as follows:
\begin{itemize}
\item We demonstrate that clustered language model embeddings can be effectively structured into temporal ordered stochastic representations via a simple multi-layer architecture.
\item We propose a novel likelihood-based metric (BBScoreV2) to evaluate temporal and structural dependencies within the stochastic representation with solid theoretical foundation.
\item We hypothesized and validated that temporal and structural information encoded in the stochastic representation, as measured by the BBScoreV2, can potentially serve as an effective and flexible metric for multiple downstream tasks such as coherence evaluate and AI-generated text detection.
\end{itemize}

\section{Related work}
Stochastic processes have demonstrated robust capabilities in modeling complex tasks across various fields, including biology \citep{Horne:2007aa} and finance \citep{Oksendal:2003}. Recently, the use of stochastic representations to model latent spaces has shown considerable promise in diverse applications such as time-series analysis \citep{Liu:2021} and dynamical flow modeling \citep{Albergo:2023, Albergo:2023BNF}. Notably, such methods also excel in generation tasks, including video generation \citep{Zhang:2023}, and long text generation \citep{Wang:2022}. A critical aspect of these tasks is to incorporate time-evolution into the latent representation, which requires capturing the time-dependent probability density functions embedded within real-world data. Generally, there are two approaches to tackle this challenge. One method is the likelihood-free training paradigm \citep{Durkan:2020}, exemplified by contrastive learning techniques, which have demonstrated significant effectiveness in handling high-dimensional data \citep{Oord:2018,Wang:2022,Zhang:2023}. This approach enables the learning of predictive density indirectly, rather than through direct reconstruction \citep{Mathieu:2021}. The alternative method is the traditional likelihood-based approach, such as stochastic interpolants \citep{Albergo:2023, Albergo:2023BNF}, which requires the pre-definition of specific target stochastic processes. Both methods exhibit substantial potential in their respective tasks.

Coherence of articles, as defined by \cite{reinhart1980conditions}, referring to the logical flow and connection of ideas in a text, is one of the most complex temporal dynamic encoded in the articles. Studies have shown that transformers, while effective in generating tasks, often struggle with capturing coherence \citep{Deng:2022}. To improve how language models learn long-text dynamics, methods using latent spaces have been developed \citep{Bowman:2016, Gao:2021}, focusing on sentence embeddings by considering neighboring utterances. However, these methods often produce static representations and neglect the text's dynamic nature. A recent approach using stochastic representations, such as the BB, incorporates "temporal dynamics" to improve long-range text dependencies \citep{Wang:2022}. This method shows promise in generating coherent long texts through capturing structural and temporal information.

In addition to generative tasks, evaluating coherence in a given text also remains a challenge \citep{Sheng:2024, Aviya2023novelcoh}. Building on stochastic concepts, \cite{Sheng:2024} developed a heuristic metric for coherence assessment, grounded in the unsupervised learning approach proposed by \citet{Wang:2022}. This score demonstrated considerable performance on artificial shuffle tasks. However, their method relies on a heuristic understanding of the BB and fails to adequately establish a theoretical foundation for the metric setup, which limit the effectiveness and flexibility of their score, particularly its sensitivity to article length.
\section{Method}
\label{sec:method}
\subsection{Brownian bridge process}
\label{sec:sr}

In this section, we introduce a stochastic representation of the encoded sequences by modeling them using BBs. We begin by defining a standard BB $\{B(t):t \in [0,T]\}$ with $B(0)=0$ and $B(T)=0$. For any $t \in [0,T]$, the process $B(t)$ follows a normal distribution $B(t) \sim N(0,t(T-t)/T)$. Additionally, for $s,t \in [0,T]$ with $s<t$, the covariance between $B(s)$ and $B(t)$ is given by $\Cov(B(s),B(t)) = s(T-t)/T$. A more general BB start from $a$ and end at $b$ can then be constructed as $a+(t/T)(b-a)+\sigma B(t)$, where $a$ and $b$ are fixed start and end points, respectively, and $\sigma$ is the standard deviation of the process.

\subsection{Contrastive learning encoder}
\label{sec:encoder_cl}
The encoder architecture consists of two components: a frozen, pre-trained language model and a trainable multilayer perceptron (MLP) network. We extract the hidden state corresponding to the end-of-sentence (EOS) token from the last layer of the language model. This hidden state serves as an input to a four-layer MLP, which is trained to map the input to the latent space. The purpose of the encoder is to learn a non-linear mapping from the raw input space to the latent space, denoted as $f_\theta: \cX \to \cS$. We train the encoder using contrastive learning (CL) loss ($L_{\rm CL}$), which enhances its ability to differentiate between positive and negative samples, following the approach of \citep{Oord:2018,Wang:2022}. 

We adopt the CL encoder framework as presented by \citet{Wang:2022}. In this framework, a key structural assumption is imposed on the latent space, namely an isotropic covariance structure represented by $\Sigma = \fI_d$, where $\fI_d$ denotes the $d$-dimensional identity matrix. Consequently, for an arbitrary starting point $\fs_0$ at time $t=0$ and an ending point $\fs_T$ at time $t=T$, the marginal distribution of $\fs_t$ at time $t$ is given by Equation \ref{eq:marginal}.

\begin{figure*}[!t]
\centering
\begin{tcolorbox}[colback=yellow!5!white, colframe=white!60!black]
\begin{alignat}{2}
&\textbf{Marginal distribution of $\fs_t$} \quad &&\fs_t \mid \fs_0, \fs_T \sim N\left(\left(1-t/T\right)\fs_0 + (t/T) \fs_T, [t(T-t)/T] \fI_d\right).
\label{eq:marginal} \\
&\textbf{Encoder contrastive loss} \quad &&L_{\rm CL} = \E\left[ - \log\frac{\exp(d(\fx_0,\fx_t,\fx_T;f_\theta))}{\sum_{(\fx_0,\fx_{t'},\fx_T)\in B} \exp(d(\fx_0,\fx_{t'},\fx_T;f_\theta)) }\right], \label{eq:cl_loss} \\
& && d(\fx_0,\fx_t,\fx_T;f_\theta) = -\frac{\norm{Tf_\theta(\fx_t) - (T-t) f_\theta(\fx_0) -t f_\theta(\fx_T)}_2^2}{2t(T-t)}. \nonumber \\
&\textbf{Log likelihood of $\Sigma$} \quad &&\ell(\Sigma|\{\bar{\fs}_i\}_{i=1}^n) =\frac{1}{2} ( d \log(2\pi) \sum_{i=1}^n (T_i-1) - d\sum_{i=1}^n \log(\lvert \Sigma_{T_i} \rvert) \label{eq:logll} \\
& &&- \log(\lvert \Sigma \rvert)\sum_{i=1}^n (T_i-1) - \sum_{i=1}^n \tr(\Sigma^{-1} (\fs_i - \fmu_i)\Sigma_{T_i}^{-1} (\fs_i - \fmu_i)^\top) ). \nonumber \\
&\textbf{Log density of $\bar{\fs}$}\quad &&\log p(\bar{\fs}|\Sigma) = -\frac{d(T-1)}{2} \log(2\pi) - \frac{d}{2} \log(\lvert \Sigma_{T} \rvert) \label{eq:density_of_s} \\
& &&- \frac{(T-1)}{2} \log(\lvert \Sigma \rvert) - \frac{1}{2} \tr(\Sigma^{-1} (\fs - \fmu)\Sigma_T^{-1} (\fs - \fmu)^\top). \nonumber\\
&\textbf{BBScoreV2 of $\bar{\fs}$}\quad &&
\cM(\bar{\fs}|\widehat\Sigma) = \log p(\bar{\fs}|\Sigma)/ [d(T-1)].
\end{alignat}
\end{tcolorbox}
\label{fig:long_equations}
\end{figure*}
Consider any triplet of observations $(\fx_1, \fx_2, \fx_3)$ with $\fx_1, \fx_2, \fx_3 \in \cX$. The goal is to ensure that $f_\theta(\fx_2)$ follows the above marginal distribution with starting point $f_\theta(\fx_1)$ and ending point $f_\theta(\fx_3)$. For a sequence of observations $(\fx_0, \ldots, \fx_T)$, let $B = \{(\fx_0, \fx_t, \fx_T)\}$ be a batch consisting of randomly sampled positive triplets $(\fx_0, \fx_t, \fx_T)$ with $0 < t < T$. Then, the CL loss function $L_{\rm CL}$ is defined by Equation \ref{eq:cl_loss}.

To further investigate the structural assumption ($\Sigma = \fI_d$) employed during encoder training, particularly given the importance of latent space correlation structure for downstream tasks, we conducted ablation studies. Specifically, we tested two different encoders: 1) CL encoder with AnInfoNCE loss: a CL loss designed by \citet{Rusak:2024:IIG} to keep learning the covariance matrix $\Sigma$ during training, and 2) a negative-log-likelihood based method (SP Encoder) which is purely based on fitting the temporal distribution of the bridge process.

\subsection{Alignment in latent space}
\label{BBScore}
To evaluate trajectories within the stochastic latent space, we propose a method to approximate the inherent correlation structure and assess both spatial and temporal properties of the encoded latents.  For an input sequence $ \bar{\fs} = (s_0,\ldots,s_T)$ with $s_t \in \mathbb R^d$ for $t=0,1,\ldots,T$, we capture temporal dependence using standard BBs. To account for structural dependence among the components, we consider $d$ independent standard BBs $B_1(t), \ldots, B_d(t)$ over the interval $[0, T]$. At each time $t$, the sequence is modeled as $s_t = \mu_t + \fW (B_1(t),\ldots,B_d(t))^\top$, where $\fW \in \bR^{d \times d}$ is a transformation matrix and $\mu_t = s_0 + (t/T) (s_{T}-s_0)$ represents the mean at time $t$. The structural dependence is captured by $\Sigma = \fW\fW^\top$. Let $\fs = (s_1,\ldots,s_{T-1})$ denote the sequence excluding the start and end points, and let $\fmu = (\mu_1,\ldots,\mu_{T-1})$ be the corresponding means.

The proposed BBScoreV2 is based on the likelihood function of the input sequences, with $\Sigma$ being the only unknown parameter. The following proposition presents the likelihood function. For the detailed proof, please check Appendix \ref{app:proof}

\begin{proposition}\label{thm:1}
Let $\Sigma_{T} \in \bR^{(T-1)\times (T-1)}$ be the covariance matrix with entries $[\Sigma_{T}]_{s,t} = s(T-t)/T$. 
For $n$ independent input sequences $\bar\fs_1,\ldots,\bar\fs_n$ with lengths $T_1+1,\ldots,T_n+1$, generated by the same $\fW$ (or equivalently, $\Sigma$), and then the log-likelihood function is defined in Equation \ref{eq:logll}.
\end{proposition}

By Proposition~\ref{thm:1}, given the input sequences,
the maximum likelihood estimate (MLE) of $\Sigma$ is.
\begin{proposition}\label{thm:2}
Under the setting of Proposition~\ref{thm:1}, 
the MLE of $\Sigma$ given $\{\bar{\fs}_i\}_{i=1}^n$ is
\begin{equation*}
    \resizebox{\columnwidth}{!}{$
\begin{aligned}
    \widehat\Sigma = \Big(\sum_{i=1}^n (T_i-1)\Big)^{-1} \Big(\sum_{i=1}^n (\fs_i - \fmu_i)\Sigma_{T_i}^{-1} (\fs_i - \fmu_i)^\top\Big).  
\end{aligned}
$}
\end{equation*}

\end{proposition}

The definition of the BBScoreV2 is therefore derived from the MLE of $\Sigma$. Consider the sequence $\bar{\fs} = (s_0, \ldots, s_T)$, with $\fs$ and $\fmu$ defined as before. To evaluate the coherence of the sequence from a domain with unknown parameters $\Sigma$, a natural approach is to compute its density under the assumed model. If $\bar{\fs}$ is a BB with covariance $\Sigma$, then by Proposition~\ref{thm:1}, the log density of $\bar{\fs}$ is given by Equation \ref{eq:density_of_s}.
To remove the length sensitive term in the log density, we design a standardized score for practical purposes, and define the score as following:
\begin{definition}[BBScoreV2]
Let $\widehat{\Sigma}$ be the estimate of $\Sigma$ from Proposition~\ref{thm:2}. The metric BBScoreV2 is defined as
\[
\cM(\bar{\fs}|\widehat\Sigma) = \log p(\bar{\fs}|\Sigma)/ [d(T-1)].
\]
\end{definition}
Given an accurate estimate $\widehat\Sigma$ of the true covariance $\Sigma$, and assuming the input sequence $\bar{\fs}$ originates from a BB process with covariance $\Sigma$, a lower BBScoreV2 value signifies a decreased likelihood of $\bar{\fs}$ being generated under $\Sigma$. Conversely, if the representation encodes a better temporal and structural information, such as encoded from a more coherent article, the probability density will be higher, resulting in a larger BBScoreV2.



In a summary, BBScoreV2 is novel in two key aspects. First, by utilizing the temporal covariance matrix $\Sigma_T$, the BBScoreV2 captures the time-dependent structure inherent in the sequence, which is essential for accurately assessing sequence temporal property, such as coherence. Second, the inclusion of the covariance matrix $\Sigma$ allows the BBScoreV2 to account for structural dependencies among the latent dimensions, providing a more comprehensive evaluation of the sequence's adherence to the assumed stochastic process.
\section{Experiments and Problems}
To understand the spatial and temporal information encoded in stochastic representations, we experimentally designed latent space visualization experiments. Subsequently, we evaluate BBScoreV2 to demonstrate its utility in downstream tasks that leverage this encoded information. Our experiments are designed to address the following three key research questions (Q):
\begin{itemize}

\item \textbf{Q1}: \textit{How is stochastic representation learning achieved, and what makes it effective?} In Section \ref{sec:latent_analysis}, we analyze the spatial structure of the latent space. 

\item  \textbf{Q2}: \textit{Can BBScoreV2 capture correct temporal information and assess document coherence?} In Section \ref{sec:coherence_eval_tasks}, we examine its performance on standard shuffle tasks (indicative of temporal understanding) and also comparing the coherence of articles of varying lengths—an evaluation that current state-of-the-art methods often cannot perform effectively.

\item \textbf{Q3}: \textit{Can we use BBScoreV2 to detect AI-generated text from human-written ones ?} In Section \ref{sec:human_AI_comparison}, we explore whether BBScoreV2 can effectively distinguish between human-written text and text generated by AI. We also compare its performance with other baselines.
\end{itemize}
To validate the above question, we design the following experiments. Moreover, in Section \ref{sec:dataset_app}, we describes the dataset utilized in these experiments and how we construct the input. 

\paragraph{Global discrimination.} We employed the Shuffle Test \citep{Barzilay:2008, Moon:2019} to assess BBScoreV2's ability to evaluate temporal information and discriminate global coherence. It involves randomly permuting sentences within a document to create an incoherent version, which is then compared against the original. Specifically, for each article, we generated 20 unique shuffled copies by permuting entire sentence blocks of varying sizes (1, 2, 5, and 10 sentences).

\paragraph{Mixed Shuffled test.} Building upon the standard Shuffle Test, we introduced a more challenging variant called the Mixed Shuffle Test. In this setup, BBScoreV2 of an original (unshuffled) article is compared against BBScoreV2 of shuffled articles drawn from the entire dataset, rather than solely against its own shuffled versions. A robust and general-purpose scoring mechanism should consistently identify the original, unshuffled article as more coherent in these broader comparisons.

\paragraph{Human-AI text discrimination.} We leverage the \text{HC3} Q\&A dataset \citep{Guo:2023} to train the encoder exclusively on human-generated answers, and subsequently apply it to unseen Q\&A pairs generated by both humans and ChatGPT. After deriving the stochastic representations, we compute the BBScoreV2 for each Q\&A pair. We evaluate multiple encoder backbones to examine the impact of the raw embeddings. Additionally, we train an encoder on the \text{WikiSection} dataset and evaluate it using the Wikipedia subset of \text{HC3}. Experiments are conducted under both the full Q\&A and answer-only settings to determine if the BBScoreV2 can effectively discriminate between ChatGPT-generated and human-written texts.

\section{Results}
\subsection{Latent space structure analysis}
\label{sec:latent_analysis}
\begin{figure*}[ht!]
\centering
\includegraphics[width=\linewidth]{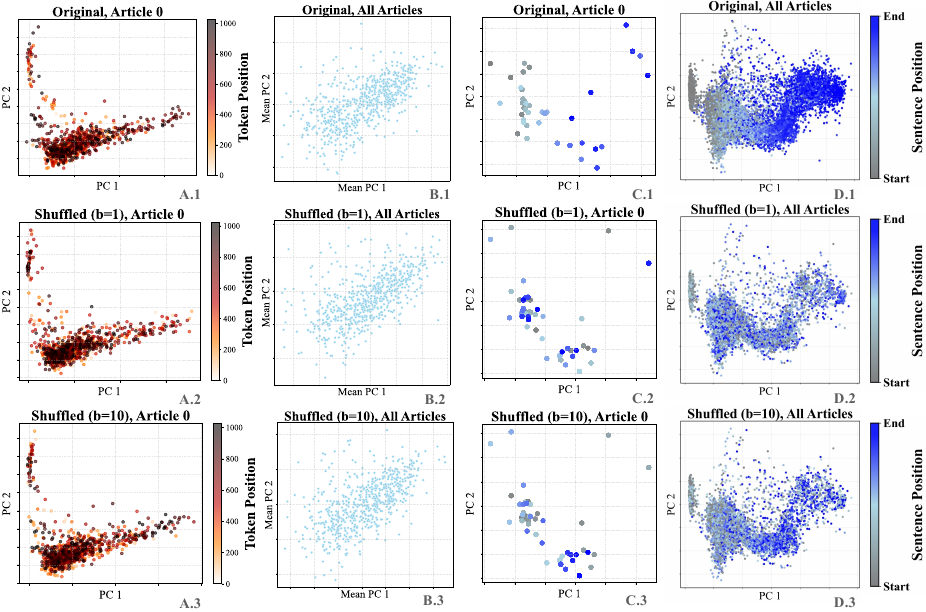}
\caption{PCA analysis of raw LM embeddings and CL encoder representations.
(A.1) Projection of raw LM embeddings for an unshuffled article onto the first two PCs. The color gradient, from light to dark red, indicates the sequential position of each token within the article. (A.2, A.3)  raw LM embeddings for shuffled versions of the same article. (B) Mean PC1 and PC2 values for all articles are plotted, with each article represented by a dot. (C.1) Latent representations from the CL encoder for an unshuffled article, where the color gradient (light to dark blue) signifies the component's position in the article sequence. (C.2, C.3) Latent patterns observed in shuffled versions of the same article. (D) Visualization of the latent trajectories for all articles.} 
\label{fig:Cluster_analysis}
\end{figure*}

Theoretically, transformer-based LLMs are argued to map articles to a latent representation that tends to form clusters. The structural properties of these clusters are believed to reflect underlying similarities and properties present in the original articles \cite{Geshkovski:2023}. 

Experimentally, we also find such clustered property. We first visualized the raw embeddings of each article from the frozen GPT-2 model. In Figure \ref{fig:Cluster_analysis} (A.1), these embeddings are projected onto their joint first two principal components (PCs), derived using PCA computed from the latents of unshuffled articles. The color gradient, from light to dark red, represents the token's sequential position within the article, from beginning to end. Notably, as shown in (A.2) and (A.3) for shuffled versions of an article, the distinct clustering property persists. This persistence, despite the disruption of sequential order, suggests that these raw LM embeddings do not clearly and inherently encode temporal information. To further substantiate this, Figure \ref{fig:Cluster_analysis} (B) plots the mean values of the first two PCs for the embeddings of each article, illustrating a tendency for articles from the same dataset to cluster together based on their raw LM embeddings. 

Subsequently, to determine the information learned by the MLP layers in our CL encoder, we analyzed its outputted stochastic representations. Figure \ref{fig:Cluster_analysis} (C.1) displays these MLP-processed latents projected via PCA. Here, a color gradient from light to dark blue indicates the component's position within the article's sequence (from begin to the end). This visualization reveals a clear temporal progression in the latent space for the original, unshuffled article. In stark contrast, Figures (C.2) and (C.3), which depict shuffled versions of the same article, demonstrate that the CL encoder's representations clearly reflect this violation of temporal order; the clear sequential pattern observed in (C.1) is visibly disrupted. Furthermore, Figure \ref{fig:Cluster_analysis} (D) presents the projection of latent trajectories for all articles. This visualization further validates our assertion that the CL encoder effectively learns and represents temporal sequence information, unlike the raw LLM embeddings.

Based on these findings, we show that the CL encoder effectively encodes temporal information into the representation. Furthermore, by evaluating the temporal structure, we can infer properties of the original articles—such as coherence—which are quantified by BBScore$^+$ and will be systematically discussed in the following sections.
\subsection{Article coherence evaluation}
\label{sec:coherence_eval_tasks}

\begin{table*}[hbtp]
\centering
\resizebox{\linewidth}{!}{%
\renewcommand{\arraystretch}{1.2}
\begin{tabular}{l|cccc|cccc}
\toprule
\multirow{2}{*}{\textbf{Methods}}  & \multicolumn{4}{c|}{\textbf{Acc. (Shuffle Task)}}  & \multicolumn{4}{c}{\textbf{Acc. (Mixed Shuffle Task)}} \\
& $\mathcal{D}_{b=1}$ & $\mathcal{D}_{b=2}$ & $\mathcal{D}_{b=5}$ & $\mathcal{D}_{b=10}$ &$\mathcal{D}_{b=1}$ & $\mathcal{D}_{b=2}$ & $\mathcal{D}_{b=5}$ & $\mathcal{D}_{b=10}$\\
\midrule
\textsc{Entity\ Grid} \citep{Barzilay:2008}  &85.73  &82.79 &75.81  &64.65  &46.10 &52.29 &53.69 &63.02 \\
\textsc{Unified Coherence} \citep{Moon:2019} & \textbf{99.73} & \text{97.86} & \text{96.90} & \text{96.09} &-- &-- &-- &-- \\
\textsc{BBScore} \citep{Sheng:2024}  & 83.39 & 80.71 & 79.36 & 78.66 &22.37 &24.94 &23.84 &19.69 \\
\cmidrule(lr){1-9}
\textsc{BBScoreV2 (GPT2-124M)}  &99.03  &98.11  &98.02  &98.17 &94.78 &\textbf{89.24} &\textbf{79.64} &70.83\\
\textsc{BBScoreV2 (LLaMA3-1B)}  &99.16 &98.37 &97.99 &97.87  &94.53 &87.86 &76.95 &\textbf{71.13} \\
\textsc{BBScoreV2 (LLaMA3-3B)}  &\text{99.57} &\textbf{98.74} &\textbf{98.14} &\textbf{98.74} &\textbf{94.97} &86.34 &73.88 &68.87 \\
\bottomrule
\end{tabular}%
}
\caption{Results of Global shuffle tasks on WikiSection. $\mathcal{D}_{b=i}$, $i=1,2,5,10$ refers to datasets constructed with varying levels of block shuffling.}
\vspace{-4mm}
\label{tab:wiki-global}
\end{table*}

As shown in Tables \ref{tab:wiki-global}, we first implement global discrimination tasks on \text{WikiSection}. In this task, BBScoreV2 significantly outperforms the BBScore and SOTA results. (See Appendix \ref{otherscore} for more details on methods we compared to.) The SOTA method, developed using a complex network structure and trained on unshuffle-shuffle data pairs, serves as a robust baseline. Our results demonstrate that BBScoreV2 surpasses the SOTA method in global discrimination tasks with larger block sizes, underscoring its potential to capture more globalized temporal properties. 

In shuffle tasks, most current high-performance methods, including the SOTA approach, rely on pairwise training and are unable to effectively compare articles of different lengths, as these models are typically constructed based on sentence-wise matching and comparisons. However, in the Mixed Shuffle test which evaluate the metric robustness across different articles, as shown in Table \ref{tab:wiki-global}, BBScoreV2 surpasses these SOTA method by generating a metric that can be compared across different articles. We use the basic entity-grid method \citep{Barzilay:2008} as a baseline and the result highlights that our score enables article-wise comparison. It also demonstrates significant potential in more complex tasks. Additionally, BBScoreV2 outperforms the BBScore in this article-wise comparison, underscoring a key contribution of our design—mitigating the effect of article length on score evaluation. This property allows for a more general comparison across diverse articles. 

We also explore the effect of different LLM backbones. We tested our model using LLaMA3-1B and LLaMA3-3B, with GPT2-124M which is the LLM model used in the main section. As summarized in Table \ref{tab:wiki-global}, we find: 1) In global shuffle task, LLaMA3-3B outperforms both GPT2-124M and the SOTA method, demonstrating its effectiveness in capturing global sequence structure; 2)  In Mixed Shuffle Task, LLaMA3-3B surpasses GPT2-124M for smaller blocks (b=1), but its performance decreases for larger blocks (b=2, b=5, b=10). This suggests a trade-off where larger models excel at capturing local details (b=1) but might sacrifice robustness for global structures (b=10). This insight highlights an intriguing direction for future exploration — different LMs may facilitate learning stochastic representations in task-specific ways.

Moreover, we evaluate the robustness of the encoded stochastic representation on a broader dataset. As shown in Table \ref{wiki-global-outofdomain}, we train the encoder on WikiText and evaluate it on WikiSection (see Appendix~\ref{sec:dataset_app} for details and comparisons about the datasets). The results indicate that our method remains highly robust in this O.O.D. setting, suggesting that the structural and temporal information captured by our model reflects fundamental patterns that generalize across different datasets.

\begin{table}[ht!]
\centering
\resizebox{\linewidth}{!}{%
\renewcommand{\arraystretch}{1.2}
\begin{tabular}{l|llll}
\toprule
\multicolumn{1}{c|}{\textbf{Methods}} & \multicolumn{4}{c}{\textbf{Shuffle Test tasks (O.O.D.)}} \\
 & $\mathcal{D}_{b=1}$ & $\mathcal{D}_{b=2}$ & $\mathcal{D}_{b=5}$ & $\mathcal{D}_{b=10}$\\
\midrule
\textsc{Unified Coherence}  &60.02 &9.63 &44.80 &66.51\\
\textsc{BBScore}  &70.32 &72.09 &76.84 &77.73\\
\rowcolor{gray!10} \textsc{BBScoreV2 }    &\textbf{91.30 } &\textbf{87.22}  &\textbf{86.14} &\textbf{88.18}\\

\bottomrule

\end{tabular}%
}
\caption{O.O.D. Task. Encoder was trained on the WikiText and evaluated on Shuffle Test tasks using the same WikiSection data to assess their performance.}
\label{wiki-global-outofdomain}
\end{table}
\vspace{-3mm}
\subsection{Human-AI discrimination tasks}
\label{sec:human_AI_comparison}

\begin{figure}[htbp]
\centering
\includegraphics[width=\linewidth]{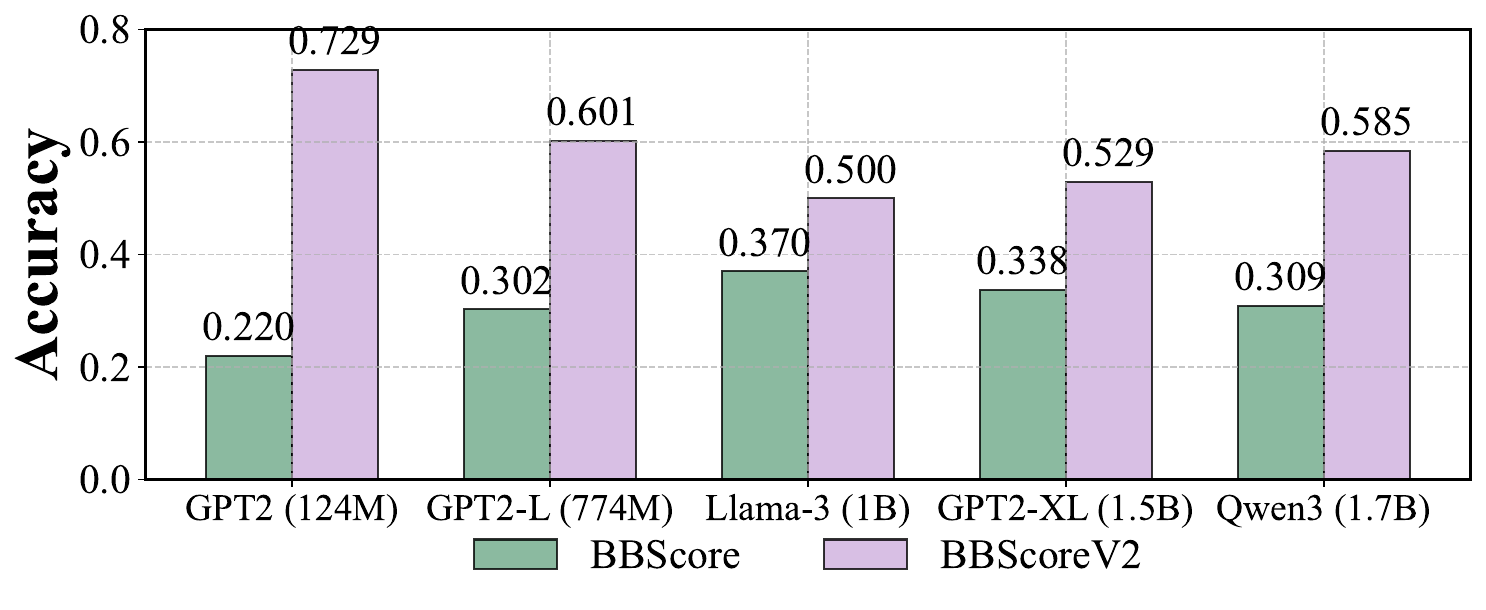}
\caption{Compare different LM backbones.}
\vspace{-3mm}
\label{fig:human-ai-model}
\end{figure}

In this task, we hypothesize that human writing, compared to AI-generated text, displays temporal dynamics and structural patterns similar to those observed in other human-written articles. Specifically, we propose that an encoder trained on a human-written dataset will more accurately capture the characteristics of human writing than those of AI-generated text, resulting in a higher likelihood for human-authored content. As shown in Figure~\ref{fig:human-ai-model}, BBScoreV2 consistently outperforms BBScore across all experimental settings. Notably, GPT2 (124M) surpasses the larger backbone models, suggesting that the quality of the learned stochastic representation does not necessarily improve with increased model size.Instead, it is the MLP module that plays the central role in shaping the stochastic representation. Among the models evaluated, GPT2 features a smaller hidden dimension of $768$, whereas both LLaMA3 and Qwen3 utilize larger hidden dimensions of $2048$. This variation suggests that the hidden dimension of the backbone model may influence performance on this task. 

Next, we use a \text{WikiSection} trained encoder to detect ChatGPT-generated answer in the Wikipedia subset of HC3. The results are shown in Table~\ref{tab:combined_human_ai_comparison} under both Q\&A and answer-only settings. To further highlight the flexibility and competitive performance of BBScoreV2 compared to LLM-based models, we assessed the perturbation discrepancy metric proposed in DetectGPT \citep{mitchell2023detectgpt}, which has high performance in AI detection tasks. Our results reveal that BBScoreV2 surpasses DetectGPT when using a comparable number of model inferences. DetectGPT's performance is influenced by a hyperparameter—the number of perturbations—which directly affects both the number of model inferences and the computational complexity. As shown in Table \ref{tab:combined_human_ai_comparison_app2} in the Appendix, we tested cases with 1 and 10 perturbations. With 1 perturbation, DetectGPT's accuracy was approximately 64\%, lower than BBScoreV2's 70\%, while requiring twice the number of model inferences per text. With 10 perturbations, DetectGPT's accuracy increased to  84\%, but this required 11 model inferences per text, making it significantly more computationally intensive than BBScoreV2.

\begin{table}[ht!]
\centering
\resizebox{\linewidth}{!}{%
\renewcommand{\arraystretch}{1.}
\begin{tabular}{l|c|c}
\toprule
\multicolumn{1}{c|}{\textbf{Methods}}  & \multicolumn{1}{c|}{\textbf{HC3 (w/o Q\&A)}}  & \multicolumn{1}{c}{\textbf{HC3 (w/ Q\&A)}} \\
\midrule
\textsc{BBScore}  &37.53 &31.47\\ 
\textsc{DetectGPT}  &64.30 &63.30\\
\rowcolor{gray!10}
\textsc{BBScoreV2} &\textbf{70.67} &\textbf{69.71}\\
\bottomrule
\end{tabular}%
}
\caption{Accuracy of the Human-AI discrimination task.}
\label{tab:combined_human_ai_comparison}
\end{table}

\vspace{-3mm}
\subsection{Ablation analysis on CL encoder}
As previously discussed, the CL encoder relies on a critical assumption of the independence and homogeneity among the dimensions of the encoded sequence which is $\Sigma = \fI_d$. To further examine this assumption, we employ two alternative methods: 

1) A likelihood-based encoder, \textbf{SP Encoder} (see Appendix~\ref{app:sp_encoder}) whose loss function is defined based on the likelihood of the Brownian bridge:  
\begin{equation}
\begin{aligned}
&L_{\rm NLL} = \sum_{j=1}^m \sum_{i=1}^{n_j} (T_i-1) \log(\lvert \Sigma_j \rvert) \\&+ \sum_{j=1}^m\sum_{i=1}^{n_j} \tr(\Sigma_j^{-1} (\fs_i^\theta - \fmu_i^\theta)\Sigma_{T_i}^{-1} (\fs_i^\theta - \fmu_i^\theta)^\top).
\end{aligned}
\end{equation}

2) A contrastive loss-based encoder, whose loss function is \textbf{AnInfoNCE} \citet{Rusak:2024:IIG}  which is capable of learning $\Sigma$ during training (see Appendix~\ref{sec:appendix:aninfonce} for details). The loss function
$L_{\rm AnInfoNCE}$ is defined as following:
{\small
\begin{equation*}
\begin{aligned}
 \E\left[ - \log\frac{\exp(d^*(\fx_0,\fx_t,\fx_T;f_\theta))}{\sum_{(\fx_0,\fx_{t'},\fx_T)\in B} \exp(d^*(\fx_0,\fx_{t'},\fx_T;f_\theta)) }\right],
\end{aligned}
\end{equation*}
}
where 
{
\begin{equation*}
\begin{aligned}
&d^*(\fx_0,\fx_t,\fx_T;f_\theta)\\ &= -\frac{\norm{f_\theta(\fx_t) - \frac{T-t}{T} f_\theta(\fx_0) - \frac{t}{T} f_\theta(\fx_T)}_{\hat{\Lambda}}^2} {2t(T-t)/T}.
\end{aligned}
\end{equation*}
}
and $\hat{\Lambda}$ is a trainable diagonal scaling matrix. Let $\mathbf{v} = f_\theta(\fx_t) - \frac{T-t}{T} f_\theta(\fx_0) - \frac{t}{T} f_\theta(\fx_T)$, then its corresponding norm is defined as:
{
\begin{equation*}
\begin{aligned}
\norm{\mathbf{v}}_{\hat{\Lambda}}^2 = \mathbf{v}^\mathrm{T} \cdot \hat{\Lambda} \cdot \mathbf{v}
\end{aligned}
\end{equation*}
}

As shown in Table~\ref{tab:aninfonce_infonce}, neither the likelihood-based nor the CL encoder with AnInfoNCE loss yields significant improvements in the shuffle test. This suggests that the MLP layers do not capture meaningful structural correlations across latent dimensions—a phenomenon also noted by \citet{Wang:2022}—and instead primarily reconstruct temporal information, which validate our assumptions on CL encoder training. The lack of performance improvement may also suggest that the pertinent correlation structure is likely inherent either within the statistical properties of the article domain or already captured within the high-dimensional embedding space of the pre-trained language model as we seen in the cluster analysis in Fig \ref{fig:Cluster_analysis}. 

\begin{table}[htbp]
\centering
\label{tab:results-comparison}
\resizebox{\linewidth}{!}{
\begin{tabular}{l|rrrr}
\toprule
\textbf{Loss Type} & $\mathcal{D}_{b=1}$ & $\mathcal{D}_{b=2}$ & $\mathcal{D}_{b=5}$ & $\mathcal{D}_{b=10}$ \\
\midrule
\textsc{AnInfoNCE} &94.63  &91.05  &92.13 & 91.70 \\
\textsc{Likelihood}  &94.42 &92.90 &90.77  &86.69\\
\rowcolor{gray!10}
\textsc{Ours} &99.03  &98.11  &98.02  &98.17\\
\bottomrule
\end{tabular}
}
\caption{Comparison of Model Performance with different loss function on WikiSection Dataset. }
\label{tab:aninfonce_infonce}
\end{table}
\vspace{-4mm}

\subsection{Computation efficiency analysis}
\begin{figure}[ht!]
\centering
\includegraphics[scale=0.12]{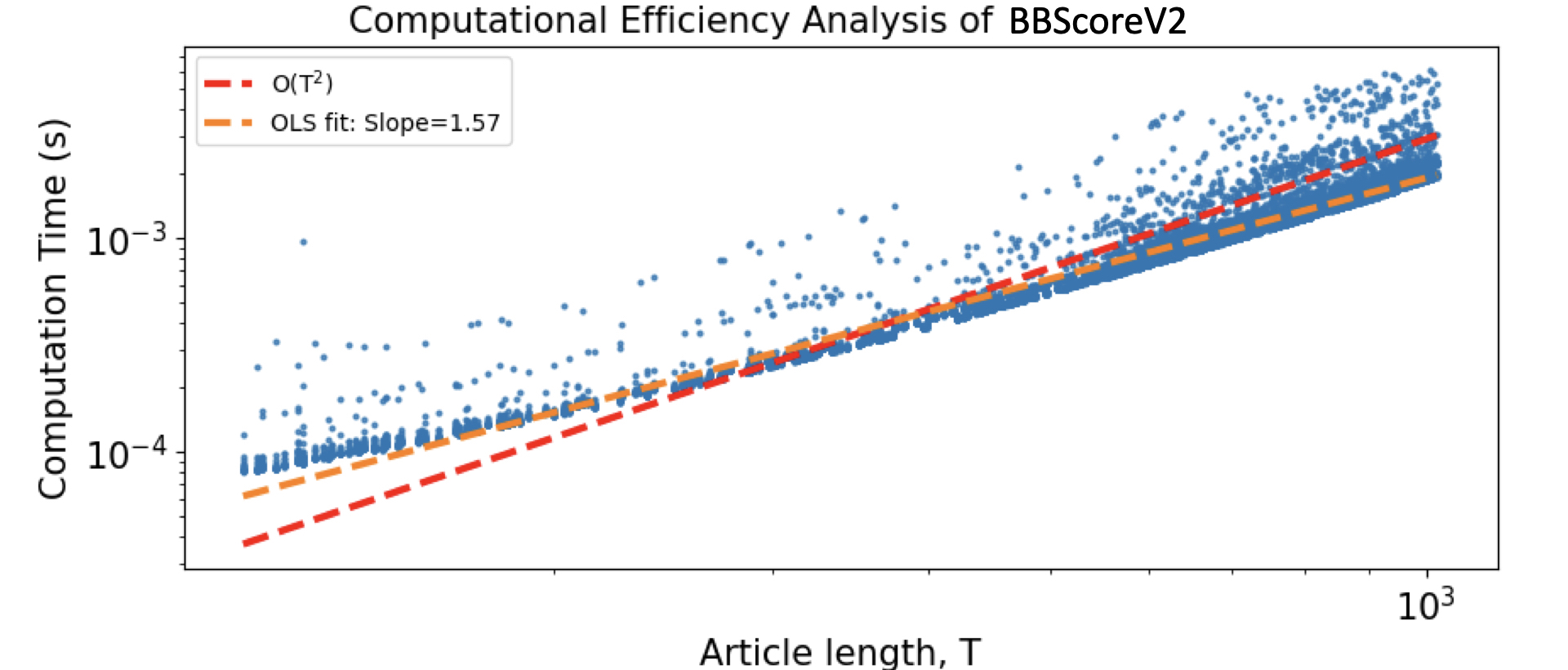}
\caption{The computation time of BBScoreV2 for different article lengths. It reveals a quadratic relationship (experimentally 1.57, theoretically 2) between article length and computation time, with each article processed in approximately $\sim 10^{-3}$ seconds.}
\label{fig:BBScore_time_analysis}
\end{figure}

We specifically analyze the computation efficiency of BBScoreV2, as shown in Figure \ref{fig:BBScore_time_analysis}, the y-axis represents computation time, while the x-axis indicates article length. The theoretical computational complexity of BBScoreV2 is \( O(T^2) \), primarily due to matrix multiplications inherent in its definition. This complexity is fundamental to fully leveraging temporal information for sequence evaluation. Empirically, the observed computation time is slightly better than the theoretical prediction, thanks to the computational acceleration. These results demonstrate that BBScoreV2 is not only feasible for real-time applications but also retains its robust evaluation capabilities.

\section{Conclusion}
In this paper, we present both a theoretical and empirical investigation into the structural and temporal properties encoded in stochastic representations of latent trajectories for NLP tasks. We analyze and visualize these properties, and introduce BBScoreV2—a novel, length-invariant metric designed to quantify such information. First we present the learned representations recovers the time dependency of the input sequences. Then validated through shuffled and mixed-shuffle tests, we show that BBScoreV2 exhibits strong performance in capturing temporal structure and generalizes effectively to out-of-distribution tasks, suggesting that these properties reflect domain-independent textual signals. Moreover, BBScoreV2 shows promising capability in distinguishing human-written from AI-generated text by leveraging encoded structural and temporal features.

Looking ahead, we aim to extend BBScoreV2 to multi-domain tasks such as domain identification, and to exploit its length-insensitive nature to develop generative models that maintain semantic coherence across varying sequence lengths. Its computational efficiency (see Fig.~\ref{fig:BBScore_time_analysis}) also makes it suitable for large-scale applications. Finally, inspired by \citet{Albergo:2023, Albergo:2023BNF}, we plan to explore more expressive bridge processes to further enhance the representational capacity of the latent space and enable richer downstream analysis and generation.

\section{Limitations}
First, our current study is constrained by limited computational resources and the lack of human-annotated data, which prevents us from evaluating BBScoreV2 against human preference—a key limitation in assessing its alignment with human judgment. Additionally, in the Human-AI text discrimination task, we were unable to evaluate the metric on a broader range of datasets or conduct more extensive comparisons across more baselines. These limitations suggest directions for future work involving large-scale human evaluation and broader benchmarking.


\bibliography{custom}

\appendix
\section{Appendix: SP Encoder}
\label{app:sp_encoder}
\subsection{Definition}
Consider a multi-domain problem with $m$ domains $\cX_1, \cX_2, \ldots, \cX_m$ each associated with domain-specific true structural parameters $\Sigma_1,\Sigma_2,\ldots,\Sigma_m$, respectively. For each domain $\cX_j$, we have $n_j$ independent raw inputs $\fx_{j1}, \ldots, \fx_{jn_j}$. We define the encoded sequences as $\bar\fs_{ji}^\theta = f_\theta(\fx_{ji})$ for $j = 1,\ldots,m$ and $i = 1,\ldots,n_j$, where $f_\theta$ is the encoder parameterized by $\theta$. When the encoder parameters reach their optimal values $\theta^*$, the sequences $[\bar{\fs}_{ji}^{\theta^*}]_{i=1}^{n_j}$ are expected to be i.i.d. samples from BBs with parameters $\Sigma_j$ for each domain $\cX_j$.

We employ the negative log-likelihood (NLL) as the loss function to train the encoder. According to Proposition~\ref{thm:1}, for each $\theta$, the negative log-likelihood for domain $\cX_j$ depends on $\Sigma_j$ and the inputs $[\fx_{ji}]_{i=1}^{n_j}$ through the expression $\sum_{i=1}^{n_j} (T_i-1) \log(\lvert \Sigma_j \rvert) + \sum_{i=1}^{n_j} \tr(\Sigma_j^{-1} (\fs_i^\theta - \fmu_i^\theta)\Sigma_{T_i}^{-1} (\fs_i^\theta - \fmu_i^\theta)^\top)$. We consider the following training process. 

\textbf{Batch Processing: }We divide the inputs $[\fx_{ji}]_{i=1}^{n_j}$ into several batches. For each batch $\cB$, we compute the batch loss using the current estimate $\widehat\Sigma_j$ of $\Sigma_j$: $\small
\sum_{i\in\cB} \tr(\widehat\Sigma_j^{-1} (\fs_i^\theta - \fmu_i^\theta)\Sigma_{T_i}^{-1} (\fs_i^\theta - \fmu_i^\theta)^\top)$. This loss function measures how well the encoded sequences fit the assumed BB model with the current structural parameter estimate.

\textbf{Handling Large Sequences: }When the sequence lengths $T_i$ are large, computing the full loss can be computationally intensive. To address this, we randomly sample a triplet of time points $t = (t_1, t_2, t_3)$ with $1 \le t_1 < t_2 < t_3 \le T_i-1$. We extract the corresponding sub-matrices $[\fs_i^\theta]_t$ and $[\fmu_i^\theta]_t$ of size $d \times 3$ from $\fs_i^\theta$ and $\fmu_i^\theta$, respectively. Let $[\Sigma_{T_i}]_t$ be the $3 \times 3$ sub-matrix of $\Sigma_{T_i}$ corresponding to the selected time points. The loss for each $i$ in the batch becomes $\tr(\widehat\Sigma_j^{-1} ([\fs_i^\theta]_t - [\fmu_i^\theta]_t)[\Sigma_{T_i}]_t^{-1} ([\fs_i^\theta]_t - [\fmu_i^\theta]_t)^\top)$. This approach reduces computational complexity while still capturing temporal dependencies at selected time points.

\textbf{Updating Structural Parameters: }After processing all batches for $\cX_j$, we update the estimate of $\Sigma_j$ using the MLE: $\widehat\Sigma_j = [\sum_{i=1}^{n_j} (T_i-1)]^{-1} [\sum_{i=1}^{n_j} (\fs_i^\theta - \fmu_i^\theta)\Sigma_{T_i}^{-1} (\fs_i^\theta - \fmu_i^\theta)^\top]$. This update aggregates information from all sequences in the domain to refine the structural parameter estimate.

\textbf{Regularization for Stability: }To stabilize the training process, we regularize $\widehat{\Sigma}_j$ by blending it with a scaled identity matrix. We compute the average variance $\widehat\sigma^2_j$ and update $\widehat{\Sigma}_j$ as follows, using a small regularization parameter $\epsilon>0$: $\widehat\Sigma_j = (1-\epsilon)[\sum_{i=1}^{n_j} (T_i-1)]^{-1} [\sum_{i=1}^{n_j} (\fs_i^\theta - \fmu_i^\theta)\Sigma_{T_i}^{-1} (\fs_i^\theta - \fmu_i^\theta)^\top] + \epsilon \widehat\sigma^2_j \fI_d$ with $\widehat\sigma^2_j = [\sum_{i=1}^{n_j} (T_i-1)d]^{-1} [\sum_{i=1}^{n_j} \tr((\fs_i^\theta - \fmu_i^\theta)\Sigma_{T_i}^{-1} (\fs_i^\theta - \fmu_i^\theta)^\top)]$. This regularization shifts $\widehat\Sigma_j$ slightly towards isotropy, improving numerical stability during optimization.

\textbf{Total Empirical Loss Function: }After iterating over all domains, the total empirical loss function becomes
\begin{equation}
\small
\begin{aligned}
L_{\rm NLL} = &\sum_{j=1}^m \sum_{i=1}^{n_j} (T_i-1) \log(\lvert \Sigma_j \rvert) \\&+ \sum_{j=1}^m\sum_{i=1}^{n_j} \tr(\Sigma_j^{-1} (\fs_i^\theta - \fmu_i^\theta)\Sigma_{T_i}^{-1} (\fs_i^\theta - \fmu_i^\theta)^\top).
\end{aligned}
\end{equation}
Minimizing this loss over $\theta$ encourages the encoder to produce sequences that align with the assumed stochastic process model across all domains.
\subsection{Training Details}
The WikiSection SP Encoder was trained on 1 A100 GPU for about 10 hours using the training set of WikiSection for 100 epochs. We used SGD optimizer and set the learning rate to be \textit{1e-9}. The $\epsilon$ in the loss function $L_{\rm NLL}$ is chosen as \textit{1e-7}.
The WikiText SP Encoder was trained on 4 A100 GPUs for roughly 20 hours for 4 epochs with WikiText dataset. For this dataset, we trained with AdamW optimizer with learning rate \textit{1e-9} and batch size \textit{32}. The $\epsilon$ in the loss function $L_{\rm NLL}$ is chosen as \textit{1e-3}. Other hyperparameters can be accessed from the configuration file in the submitted code. 
Our empirical results show incorporating $\hat\sigma_j$ into the $\widehat\Sigma_j$ makes no significant results in the downstream tasks, thus we disregard $\hat\sigma_j$ during encoder training.

\subsection{Hyper-parameter Tuning}
While training the SP Encoder, we experimented with different $\epsilon$ in $L_{\rm NLL}$ to see its impact on the performance of the trained encoder. Note that $\epsilon$ determines the perturbation added to the matrix $\widehat\Sigma$. The eigenvalues of the initial $\widehat\Sigma$ range from $10^{-6}$ to $10^{-1}$, with the majority of which lying in $[10^{-3}, 10^{-5}]$. Thus we tested the following three different $\epsilon$:
\begin{itemize}
    \item Large $\epsilon = 10^{-3}$ that is larger that most eigenvalues of $\widehat\Sigma$.
    \item Medium $\epsilon = 10^{-5}$ that is about the same scale of most eigenvalues of $\widehat\Sigma$.
    \item Small $\epsilon = 10^{-7}$ that is smaller than most eigenvalues of $\widehat\Sigma$.
\end{itemize}
We choose the small $\epsilon$ based on the performance.

\section{Appendix: AnInfoNCE}
\label{sec:appendix:aninfonce}

Empirically, it has been observed that only one dimension of the encoder's output is effective, which is undesirable for representing complex sequences. 

As the current CL theory does not account for augmentations that affect latents to a different extent, \citet{Rusak:2024:IIG} proposed an anisotropic model for CL. Based on their work, we design the following new CL Loss $L_{\rm AnInfoNCE}$, defined as:
{\small
\begin{equation*}
\begin{aligned}
 \E\left[ - \log\frac{\exp(d^*(\fx_0,\fx_t,\fx_T;f_\theta))}{\sum_{(\fx_0,\fx_{t'},\fx_T)\in B} \exp(d^*(\fx_0,\fx_{t'},\fx_T;f_\theta)) }\right],
\end{aligned}
\end{equation*}
}
where 
{\small
\begin{equation*}
\begin{aligned}
d^*(\fx_0,\fx_t,\fx_T;f_\theta) &= -\frac{\norm{f_\theta(\fx_t) - \frac{T-t}{T} f_\theta(\fx_0) - \frac{t}{T} f_\theta(\fx_T)}_{\hat{\Lambda}}^2} {2\sigma^2},\\ \text{and}~\sigma^2 &= \frac{t(T-t)}{T}.
\end{aligned}
\end{equation*}
}
and $\hat{\Lambda}$ is a trainable diagonal scaling matrix. Let $\mathbf{v} = f_\theta(\fx_t) - \frac{T-t}{T} f_\theta(\fx_0) - \frac{t}{T} f_\theta(\fx_T)$, then its corresponding norm is defined as:
{\small
\begin{equation*}
\begin{aligned}
\norm{\mathbf{v}}_{\hat{\Lambda}}^2 = \mathbf{v}^\mathrm{T} \cdot \hat{\Lambda} \cdot \mathbf{v}
\end{aligned}
\end{equation*}
}

\section{Proof}
\label{app:proof}
\subsection{Proof of Proposition~\ref{thm:1}}

\begin{proof}
We fix the start and end points $s_0$ and $s_T$ and calculate the likelihood function of the input sequence $\fs$.

Given that $s_t - \mu_t = \fW (B_1(t),\ldots,B_d(t))^\top$, and considering the independence of $B_1(t),\ldots,B_d(t)$ along with the properties of the standard BB, we have for any $t, t' \in \{1,2,\ldots,T-1\}$: $\E[s_t-\mu_t] = 0$, $\Var[s_t] = [\Sigma_{T}]_{t,t} \Sigma$ and $\Cov[s_t,s_{t'}] = [\Sigma_{T}]_{t,t'} \Sigma$. Therefore, the vectorized form of $\fs - \fmu$ follows a multivariate normal distribution:
\begin{align*}
\vect(\fs - \fmu) \sim N(0, \Sigma_{T} \otimes \Sigma),
\end{align*}
where $\vect(\cdot)$ denotes vectorization and $\otimes$ represents the Kronecker product.

Using the likelihood function of the multivariate normal distribution, we have:
\begin{align*}
\small
L(\Sigma|\bar{\fs}) =& (2\pi)^{-d(T-1)/2} \lvert  \Sigma_{T} \otimes \Sigma \rvert^{-1/2} \\
&\cdot \exp[-\vect(\fs - \fmu)^\top [\Sigma_{T} \otimes \Sigma]^{-1}\vect(\fs - \fmu)/2].
\end{align*}
Using properties of the Kronecker product, we have $\lvert  \Sigma_{T} \otimes \Sigma \rvert = \lvert  \Sigma_{T} \rvert^d \lvert \Sigma \rvert^{T-1}$ and then
\begin{align*}
& \vect(\fs - \fmu)^\top [\Sigma_T \otimes \Sigma]^{-1}\vect(\fs - \fmu) \\
&= \vect(\fs - \fmu)^\top [\Sigma_T^{-1} \otimes \Sigma^{-1}]\vect(\fs - \fmu)\\
&= \vect(\fs - \fmu)^\top \vect(\Sigma^{-1} (\fs - \fmu)\Sigma_T^{-1})\\
&= \tr((\fs - \fmu)^\top \Sigma^{-1} (\fs - \fmu)\Sigma_T^{-1})\\
&= \tr(\Sigma^{-1} (\fs - \fmu)\Sigma_T^{-1} (\fs - \fmu)^\top).
\end{align*}
Therefore, the likelihood function becomes:
\begin{align*}
L(\Sigma|\bar{\fs}) =& (2\pi)^{-d(T-1)/2} \lvert  \Sigma_{T} \rvert^{-d/2} \lvert \Sigma \rvert^{-(T-1)/2} \\&\cdot \exp[-\tr(\Sigma^{-1} (\fs - \fmu)\Sigma_T^{-1} (\fs - \fmu)^\top)/2].
\end{align*}
Taking the logarithm, the log-likelihood function is: 
\begin{align*}
\small
\ell(\Sigma|\bar{\fs}) =& -\frac{d(T-1)}{2} \log(2\pi) - \frac{d}{2} \log(\lvert  \Sigma_{T} \rvert) \\
&- \frac{(T-1)}{2} \log(\lvert  \Sigma \rvert)\\
&- \frac{1}{2} \tr(\Sigma^{-1} (\fs - \fmu)\Sigma_T^{-1} (\fs - \fmu)^\top).
\end{align*}

For $n$ independent input sequences $\bar\fs_1,\ldots,\bar\fs_n$ with lengths $T_1+1,\ldots,T_n+1$, generated by the same $\Sigma$, the total likelihood is:
\[
L(\Sigma|\{\fs_i\}_{i=1}^n) =  \Pi_{i=1}^n L(\Sigma|\fs_i).
\]
Then the total log-likelihood function is
\begin{align*}
\small
& \ell(\Sigma|\{\fs_i\}_{i=1}^n) = \sum_{i=1}^n \ell(\Sigma|\fs_i)\\
=& -\frac{d\sum_{i=1}^n (T_i-1)}{2} \log(2\pi) - \frac{d}{2} \sum_{i=1}^n \log(\lvert  \Sigma_{T_i} \rvert) \\
&- \frac{\sum_{i=1}^n (T_i-1)}{2} \log(\lvert  \Sigma \rvert)\\
&- \frac{1}{2} \sum_{i=1}^n \tr(\Sigma^{-1} (\fs_i - \fmu_i)\Sigma_{T_i}^{-1} (\fs_i - \fmu_i)^\top).
\end{align*}
\end{proof}

\subsection{Proof of Proposition~\ref{thm:2}}

\begin{proof}
To find the MLE of $\Sigma$, we need to minimize the negative log-likelihood function, which is equivalent to minimizing:
\begin{equation}
\scriptsize
g(\Sigma) = \sum_{i=1}^n (T_i-1) \log(\lvert \Sigma \rvert) + \sum_{i=1}^n \tr(\Sigma^{-1} (\fs_i - \fmu_i)\Sigma_{T_i}^{-1} (\fs_i - \fmu_i)^\top).
\end{equation}

Since $\Sigma = \fW\fW^\top$ is positive definite, we can compute the gradient of $g(\Sigma)$ with respect to $\Sigma$. Note that:
\begin{align*}
\scriptsize
    &\frac{\dd}{\dd \Sigma} \log(\lvert \Sigma \rvert) = \Sigma^{-1},\\
    &\frac{\dd}{\dd \Sigma} \tr(\Sigma^{-1} (\fs_i - \fmu_i)\Sigma_{T_i}^{-1} (\fs_i - \fmu_i)^\top) \\
    =& -\Sigma^{-1} (\fs_i - \fmu_i)\Sigma_{T_i}^{-1} (\fs_i - \fmu_i)^\top \Sigma^{-1}.
\end{align*}

We compute the gradient:
\begin{equation}
\scriptsize
\frac{\dd}{\dd \Sigma} g(\Sigma) = \Big(\sum_{i=1}^n (T_i-1)\Big) \Sigma^{-1} - \Sigma^{-1} \Big(\sum_{i=1}^n (\fs_i - \fmu_i)\Sigma_{T_i}^{-1} (\fs_i - \fmu_i)^\top\Big) \Sigma^{-1}.
\end{equation}

Setting the gradient to zero for minimization, we have:
\begin{equation}
\scriptsize
\widehat\Sigma = \Big(\sum_{i=1}^n (T_i-1)\Big)^{-1} \Big(\sum_{i=1}^n (\fs_i - \fmu_i)\Sigma_{T_i}^{-1} (\fs_i - \fmu_i)^\top\Big).
\end{equation}
As shown, the MLE estimate for $\Sigma$ is obtained.
\end{proof}

\section{Datasets}
\label{sec:dataset_app}
\textbf{WikiSection: } We use dataset introduced in \cite{arnold-etal-2019-sector} which contains selected Wikipedia articles on the topic of global cities and have clear topic structures. Each article in this collection follows a pattern certain sections such as abstract, history, geographics and demographics. The training split contains 2165 articles and the test split has 658 articles.

\textbf{HC3: } The Human ChatGPT Comparison Corpus (HC3) \citep{Guo:2023} includes comparative responses from human experts and ChatGPT, covering questions from various fields such as open-domain, finance, medicine, law, psychology and Wikipedia. We construct the input by concatenating the \textit{Question} and \textit{Answers} together as a single document and label whether it is ChatGPT generated by the source of the answers. We also use the data without Q\&A settings and only treat the answer part as a single document.

\textbf{WikiText: } WikiText language modeling dataset \citep{merity2016pointer} is a much larger set of verified good and featured articles extracted from Wikipedia compared to WikiSection,we further compare these two dataset (Section \ref{sec:wiki_compare}) and show that there is only $\sim 1\%$ potential overlap in topics. We used \textit{WikiText-103-v1} collection in specific for experiments. This dataset encompass over 100 million tokens from 29,061 full articles. The dataset is assessible through Huggingface \footnote[1]{\url{https://huggingface.co/datasets/EleutherAI/WikiText_document_level}}. 

\paragraph{Difference between \textbf{WikiSection} and \textbf{WikiText}} \label{sec:wiki_compare}
The \textbf{WikiSection} dataset comprises 2,165 articles describing cities from Wikipedia, while \textbf{WikiText} includes 29,061 featured or high-quality articles covering a broader range of topics. The \textbf{WikiSection} dataset is most similar to the ``places" category in \textbf{WikiText}, which contains approximately 500 articles. To ensure dataset exclusivity, we used string match to check the overlapping. The regular expression query we used is \texttt{'(a|the) ([\textbackslash w\textbackslash s]*)?(city|town) in'} as it is contained in 1,721 articles out of 2,165 in WikiSection dataset. Using the same query, we examined the WikiText dataset and checked the intersection of first word of the article from both search result. After manually getting rid of false positives, there are around 30 documents found overlap in both datasets. We argue that with that amount of  ($\sim$0.1\%) contamination, \textbf{WikiSection} can be considered out of domain of \textbf{WikiText}.

\section{Other scores used in this paper}\label{otherscore}
\paragraph{Entity Grid} \citet{Barzilay:2008} is the most recognized entity-based approach. It creates a two-way contingency table for each input document to track the appearance of entities in each sentence. We use Stanford’s CoreNLP to annotate the documents and the implementation provided in the Coheoka library\footnote[2]{\url{https://github.com/kigawas/coheoka}} to obtain the Entity Grid score.
\paragraph{Unified Coherence} \citet{Moon:2019} presents a neural-based entity-grid method that integrates sentence grammar, inter-sentence coherence relations, and global coherence patterns, achieving state-of-the-art results in artificial tasks. 

\paragraph{BBScore} \citet{Sheng:2024} introduces BBScore, and also check the main text for a comprehensive comparison between BBScore and BBScoreV2. 

\section{Human-AI comparison test}
Table \ref{tab:combined_human_ai_comparison_app_bb} presents the performance of BBScoreV2 computed with different $\widehat\Sigma \in \mathbb{R}^d$, while Table \ref{tab:combined_human_ai_comparison_app_bbh} shows the performance of the BBScore  with various $\widehat\sigma \in \mathbb{R}$, where the subscript indicates the dataset used for approximation.

The clear improvement over the BBScore demonstrates that accurately capturing structural and temporal information can significantly enhance the model’s accuracy. Table \ref{tab:combined_human_ai_comparison_app2} display the performance of DetectGPT with more inferences which significantly improves its performance while also takes much longer time to infer.

\begin{table*}[ht!]
\centering
\resizebox{0.9\linewidth}{!}{%
\begin{tabular}{c|cccccc}
\hline
\multirow{2}{*}{} & \multicolumn{3}{c}{\textbf{Human AI comparison}} & \multicolumn{3}{c}{\textbf{Human AI comparison with Q\&A}} \\
& Human ($\widehat \Sigma_{human}$) & Human ($\widehat \Sigma_{ai}$) & Human ($\widehat \Sigma_{wiki}$) & Human ($\widehat \Sigma_{human}$) & Human ($\widehat \Sigma_{ai}$) & Human ($\widehat \Sigma_{wiki}$) \\
\hline
AI ($\widehat \Sigma_{human}$) & 70.07 & 70.55 &- & 69.00 & 69.60 &- \\
\hline
AI ($\widehat \Sigma_{ai}$) & 59.98 & 61.52 &- & 58.19 & 59.74 &- \\
\hline
AI ($\widehat \Sigma_{wiki}$) &- &- &\textbf{70.67} &- &- & \textbf{69.71} \\
\hline
\end{tabular}%
}
\caption{Combined accuracy of human AI comparison and human AI comparison with Q\&A}
\label{tab:combined_human_ai_comparison_app_bb}
\end{table*}

\begin{table*}[ht!]
\centering
\resizebox{0.9\linewidth}{!}{%
\begin{tabular}{c|cccccc}
\hline
\multirow{2}{*}{} & \multicolumn{3}{c}{\textbf{Human AI comparison}} & \multicolumn{3}{c}{\textbf{Human AI comparison with Q\&A}} \\
& Human ($\widehat \sigma_{human}$) & Human ($\widehat \sigma_{ai}$) & Human ($\widehat \sigma_{wiki}$) & Human ($\widehat \sigma_{human}$) & Human ($\widehat \sigma_{ai}$) & Human ($\widehat \sigma_{wiki}$) \\
\hline
AI ($\widehat \sigma_{human}$) &35.99 &\textbf{45.13} &-  &35.04  &\textbf{38.84}  &- \\
\hline
AI ($\widehat \sigma_{ai}$) &26.37 &37.05 &- &33.73 &38.12  &- \\
\hline
AI ($\widehat \sigma_{wiki}$) &- &- &37.53 &- &- &31.47 \\
\hline
\end{tabular}%
}
\caption{Human-AI Task Results with BBScore \citep{Sheng:2024}.}
\label{tab:combined_human_ai_comparison_app_bbh}
\end{table*}

\begin{table*}[ht!]
\centering
\resizebox{0.8\linewidth}{!}{%
\begin{tabular}{c|cccc}
\hline
 & \multicolumn{2}{c}{\textbf{Human AI comparison}} & \multicolumn{2}{c}{\textbf{Human AI comparison with Q\&A}} \\
 Number of Perturbations & 1 & 10 & 1 & 10 \\
\hline
 Number of LLM Inferences & \multicolumn{4}{c}{Number of Perturbations + 1}\\
Accuracy & 64.30 & 84.89 & 63.30 & 83.13 \\
\hline
\end{tabular}%
} 
\caption{Human-AI Task Results with DetectGPT \citep{mitchell2023detectgpt}. As a comparison, BBScoreV2 only requires one LLM inference.}
\label{tab:combined_human_ai_comparison_app2}
\end{table*}

\end{document}